\documentclass{ecai}
\usepackage{graphicx}
\usepackage{latexsym}
\usepackage{amsmath} 
\usepackage{amsfonts} 

\begin{document}

\begin{frontmatter}

\title{Analyzing Credit Risk Model Problems through NLP- \\
Based Clustering and Machine Learning: Insights from Validation Reports}

\author[A,C]{\fnms{Szymon}~\snm{Lis}\orcid{0000-0001-9878-3856}\thanks{Corresponding Author. Email: sm.lis@student.uw.edu.pl.}}
\author[C]{\fnms{Mariusz}~\snm{Kubkowski}\orcid{0000-0002-1453-5589}}
\author[C]{\fnms{Olimpia}~\snm{Borkowska}}
\author[B,C]{\fnms{Dobromił}~\snm{Serwa}\orcid{0000-0002-1040-181}}
\author[C]{\fnms{Jarosław}~\snm{Kurpanik}}

\address[A]{Department of Quantitative Finance, University of Warsaw}
\address[B]{The Collegium of Economic Analysis, Warsaw School of Economics}
\address[C]{ING Hubs Poland}

\begin{abstract}
This paper explores the use of clustering methods and machine learning algorithms, including Natural Language Processing (NLP), to identify and classify problems identified in credit risk models through textual information contained in validation reports. Using a unique dataset of 657 findings raised by validation teams in a large international banking group between January 2019 and December 2022. The findings are classified into nine validation dimensions and assigned a severity level by validators using their expert knowledge. The authors use embedding generation for the findings' titles and observations using four different pre-trained models, including "module\_url" from TensorFlow Hub and three models from the SentenceTransformer library, namely "all-mpnet-base-v2", "all-MiniLM-L6-v2", and "paraphrase-mpnet-base-v2". The paper uses and compares various clustering methods in grouping findings with similar characteristics, enabling the identification of common problems within each validation dimension and severity. The results of the study show that clustering is an effective approach for identifying and classifying credit risk model problems with accuracy higher than 60\%. The authors also employ machine learning algorithms, including logistic regression and XGBoost, to predict the validation dimension and its severity, achieving an accuracy of 80\% for XGBoost algorithm. Furthermore, the study identifies the top 10 words that predict a validation dimension and severity. Overall, this paper makes a contribution by demonstrating the usefulness of clustering and machine learning for analyzing textual information in validation reports, and providing insights into the types of problems encountered in the development and validation of credit risk models.
\end{abstract}

\end{frontmatter}

\section{Introduction}

The development, implementation, and maintenance of credit risk models are subject to stringent regulations (e.g. \cite{kn:BCBS2017}, \cite{kn:EBA217}, \cite{kn:EC2015}). Consequently, banking institutions employ teams of experts who exercise utmost caution in constructing high-quality models. These models undergo a comprehensive process of verification, assessment, and validation. To ensure alignment with the bank's lending and collection processes, the model structure and data undergo scrutiny and challenges from the bank's business units. Ongoing monitoring of model performance allows for the identification of necessary parameter adjustments over time.

Furthermore, an independent validation team conducts thorough analyses of all aspects of the modeling process, including verification of modeling codes and statistical results. The internal bank audit serves to verify compliance with internal policies, procedures, and risk management practices. Finally, the regulator assesses the model's fitness for its intended purposes through a comprehensive evaluation or review of its accuracy, robustness, and adherence to regulatory requirements.

The requirement to monitor, verify, and document the performance of Internal Ratings-Based (IRB) models provides a retrospective view of past issues encountered during the modeling process. These verification activities are meticulously documented in monitoring and validation reports, which hold particular significance for our study. Validation reports bear similarities to referee reports in scientific papers, as independent reviewers—model validators—assess the clarity of documentation, design of the model, and the statistical methods employed in credit risk prediction. It can be assumed, therefore, that specialists in the field of credit risk modeling are actively involved in the development and validation of such models. This raises the intriguing question of the challenges faced by model developers within this sophisticated and specialized environment.

This paper aims to investigate how textual information contained in validation reports of credit risk models can be leveraged to identify and categorize problems encountered within specific dimensions. For instance, if a credit risk model exhibits poor predictive power, as indicated by a low out-of-sample AUC (area under the curve) statistic, this would be classified as an issue within the model output dimension. Similarly, an unclear description of the modeling process would be attributed to a documentation flaw. Moreover, the presence of numerous missing values in the explanatory variables of the modeling dataset would be addressed as an issue within the model input dimension.

Our study focuses on statistical models developed under the Basel II guidelines, with a specific emphasis on the Internal Ratings-Based (IRB) approach. The IRB approach involves the use of credit risk models developed internally by banks for calculating regulatory capital. Factors motivating our choice to investigate these models include standardization and comparability of IRB models, stringent regulatory requirements affecting general quality of these models, and availability of validation reports written by independent reviewers.

The paper is structured as follows: Section 2 provides a literature review, Section 3 presents the data description and preprocessing techniques employed, Section 4 outlines the methodology, Section 5 presents the results, and finally, Section 6 concludes the paper.

\section{Literature Review}

The application of textual data in finance has a relatively short history. Early attempts by Antweiler and Frank (\cite{kn:Antweiler2004}) and Tetlock (\cite{kn:Tetlock2007}) focused on utilizing textual information, primarily through count-based approaches, for predicting stock prices based on Internet or news articles using the Naive Bayes algorithm and supervised word-counting techniques. Subsequent studies, such as those by Gentzkow and Shapiro (\cite{kn:Gentzkow2010}) and Engelberg and Parsons (\cite{kn:Engelberg2011}), expanded the scope by employing text to estimate parameters in structural models or infer causal relationships.

While these supervised learning approaches offer economic interpretability, they often require substantial domain knowledge from researchers and are computationally demanding. In recent years, advancements in Natural Language Processing (NLP) literature, particularly neural network language models, have emerged as an alternative approach. These models effectively capture syntactic and semantic structures with computational tractability, although they may lack transparency and interpretability.

Textual analysis has found emerging applications in finance and economics, encompassing studies that quantify information content in IPO prospectuses \cite{kn:Hanley2010}, FOMC announcement transcripts \cite{kn:Jegadeesh2017}, \cite{kn:Hansen2018}, firms' 10K and 10Q filings \cite{kn:Cohen2016}, patents related to FinTech innovation \cite{kn:Chen2019}, and documents capturing market sentiments \cite{kn:Loughran2011}, \cite{kn:Bollen2011} and \cite{kn:Wang2019}. These studies leverage computational linguistics tools and novel data sources to extract valuable insights. Notably, Hassan, Hollander, van Lent, and Tahoun (\cite{kn:Hassan2017}) provide an example that utilizes both computational linguistics tools and novel data.

In this context, our framework represents one of the pioneering efforts to apply clustering techniques to data from one of Europe's largest bank groups, aiming to explore the suitability of existing dimension applications and potential extensions in credit risk model validation. To achieve this objective, several clustering algorithms, including k-means, spectral clustering, birch, agglomerative clustering, and Mini Batch k-means, were employed. The iterative nature of the approach assigns each observation in the dataset to one of the k groups based on feature similarity. Alternatively, hierarchical cluster analysis starts with each observation as a separate cluster and progressively merges clusters based on feature dissimilarity until only one cluster remains. Hierarchical clustering is particularly suitable for datasets with a small number of observations, whereas k-means is more appropriate when the dataset contains a large number of observations. Generally, k-means is recommended for datasets with more than 200 observations \cite{kn:Steinley2007}.

\section{Data}

\subsection{Data description}

We utilize a unique dataset comprising descriptions of 657 findings and their corresponding characteristics. These findings were identified by validation teams during the validation process of internal ratings-based (IRB) credit risk models  within a large international banking group, spanning from January 2019 to December 2022. The models under the scope of the study are as follows: 

\begin{enumerate}
\item{Probability of Default (PD) is a credit risk model that estimates the likelihood of a borrower defaulting on its obligations within a given time frame.}
\item{Loss Given Default (LGD): LGD is a credit risk model that quantifies the expected loss a lender would incur if a borrower defaults on its obligations, considering factors such as collateral and recovery rates.}
\item{Exposure at Default (EAD): EAD is a credit risk model that calculates the amount of exposure a lender has to a borrower at the time of default, incorporating both on-balance-sheet and off-balance-sheet exposures.}
\end{enumerate}

In the life-cycle of credit risk models, regular verification is conducted to assess the quality and performance of these models. Newly developed models undergo a comprehensive pre-approval validation, which entails thorough checks across all model dimensions, including model documentation, design, use, and implementation. Subsequently, during the obligatory annual periodical validation activities, the focus mainly shifts to the model input, model environment, and model output dimensions. However, it is important to note that findings related to any validation dimension can be raised by experts at any time, not limited to pre-approval or periodical validations.

Each individual finding record within our dataset contains a title (providing a concise description of the finding), a description (explaining the issues identified in the investigated model), the finding date (indicating when the finding was raised), the person responsible for addressing the finding (person to act), the action plan (planned activities to rectify the issue), and the due date (mandatory deadline for resolving the finding).
A crucial element associated with each finding is the assigned validation dimension, which identifies the specific area where a problem has been identified and where targeted actions must be undertaken to address the issue. Our dataset distinguishes nine validation dimensions:

\begin{enumerate}
\item{Documentation: Pertaining to the description of the model, encompassing technical model documentation, implementation specification documents, supplementary documents, modeling codes, and datasets.}
\item{Model input: Encompassing all data used in the modeling process, including aspects related to data quality, availability, adequacy, and validity.}
\item{Model environment: Focused on ensuring the alignment of the developed model with its intended application scope, including data representing the model scope, internal processes, and external environment \cite{kn:EBA217}. This dimension shares similarities with the model input dimension due to its relationship with data.}
\item{Model output: Concerned with the quality of predictions generated by the model, encompassing analyses of predictive power, estimation bias, calibration quality, and other relevant measures.}
\item{Model design: Covers the statistical specification of the model, applied estimation and calibration methodologies, statistical tests used to assess model adequacy, and related topics.}
\item{Impact assessment: Involves calculating the economic impact of the model output, such as predicted credit risk, on a bank's financial situation. It also includes a description of the main factors influencing the calculated impact.}
\item{Margin of conservatism: a technical dimension associated with the appropriate calculation of an additional level of conservatism beyond the best estimate of predicted credit risk in a credit risk model. This accounts for uncertainties that may introduce bias to the risk quantification.}
\item{Model use: Considers the various applications of the model in banking activities and assesses its suitability for all intended purposes.}
\item{Model implementation: Covers the proper implementation of the developed model within the banking information technology systems.}
\end{enumerate}

Figure 1 presents the distribution of "Model Category" and "Dimension" variables within our dataset. Each bar on the plot represents a unique combination, and its height corresponds to the count of occurrences for that specific combination.

\begin{figure}
\centering
\includegraphics[width=0.5\textwidth]{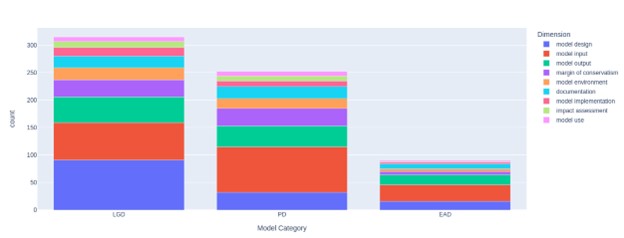}
\caption{The distribution of "Model Category" and "Dimension" variables}
\label{procstructfig}
\end{figure}

By examining these validation dimensions, we aim to explore the appropriateness of their current application and identify potential extensions for model validation in credit risk. To achieve this goal, we employed several cluster analysis algorithms, including k-means, spectral clustering, birch, agglomerative clustering, and Mini Batch k-means. The choice of algorithm depends on the size of the dataset, with hierarchical clustering suitable for small datasets and k-means recommended for larger datasets \cite{kn:Steinley2007}.

\subsection{Data transformation}

The data transformation process is crucial for utilizing the descriptions and titles of findings as predictors in our models. To prepare the textual data, we employ several techniques \cite{kn:Mooney2000}. Initially, we remove stop words, punctuation marks, and symbols from both the descriptions and titles. This step helps eliminate commonly used words and non-essential characters that do not contribute significantly to the predictive power.

Next, we convert the processed texts into sequences of lexical tokens, typically single words. This allows us to break down the textual information into individual units and analyze their significance. Additionally, we apply normalization and lemmatization techniques to the lexical tokens. Normalization ensures that variations in word forms (e.g., plural and singular) are standardized, reducing the number of unique expressions. Lemmatization further aids in reducing word variations by converting them to their base or root form. By applying these methods, we can identify specific expressions that are valuable for predicting the validation dimensions.

To enable the utilization of textual information in our prediction models, we transform the finding titles and descriptions into numerical vectors. We accomplish this through the utilization of phrase embedding, a technique where the meaning of phrases is encoded into real-valued vectors. These vectors capture the semantic relationships between phrases, enabling vectors with similar values to be indicative of similar meanings. To perform this transformation, we leverage three widely used pre-trained general-purpose sentence-transformer models: "all-mpnet-base-v2," "all-MiniLM-L6-v2," and "paraphrase-mpnet-base-v2." These models provide effective representations of the textual data, facilitating the integration of the findings' descriptions and titles into our predictive models.

\section{Methodology}

\subsection{Clustering}

In our study, we employed five clustering algorithms to perform data clustering: k-means, spectral clustering, Birch, agglomerative clustering, and Mini Batch k-means. Each algorithm possesses unique characteristics and is suitable for different types of datasets. Here, we provide a concise overview of each algorithm and outline how they were implemented in our study.

K-means, introduced by Linde et al. (\cite{kn:Linde1980}), is a widely used unsupervised machine learning algorithm for cluster analysis. Its objective is to partition a given dataset into k clusters based on the similarity of the data points. 

Spectral clustering, popularized by Shi and Malik (\cite{kn:Shi2000}) and Jordan and Weiss (\cite{kn:Ng2001}), utilizes the spectral theory of graphs to identify clusters. It performs a spectral decomposition of a similarity matrix derived from the dataset to uncover meaningful clusters.

BIRCH (Balanced Iterative Reducing and Clustering using Hierarchies), introduced by Zhang et al. (\cite{kn:Zhai2016}), is a hierarchical clustering algorithm designed to handle large datasets with numerous dimensions. It is a memory-efficient algorithm capable of incrementally clustering the dataset, making it suitable for online and real-time clustering applications. 

Agglomerative clustering, proposed by Sneath and Sokal (\cite{kn:Sneath1973}), is a hierarchical clustering algorithm that starts with each data point as a separate cluster and iteratively merges the two closest clusters until the desired number of clusters is achieved. In our study, we utilized Ward's linkage as the linkage criterion, where the distance between two clusters is measured by the increase in the sum of squared distances within the clusters upon merging \cite{kn:Ward1963}.

Mini Batch k-means, introduced by Sculley (\cite{kn:Sculley2010}), is a variant of the k-means clustering algorithm that employs mini-batches of data to update the cluster centroids. This makes it more memory-efficient and faster compared to the standard k-means algorithm.

The selection of these clustering algorithms was driven by the specific characteristics of our dataset and the research objectives of our study. By utilizing a diverse set of algorithms, we aimed to explore different clustering perspectives and ensure the robustness of our findings. Additionally, these algorithms are well-established in the field and have been extensively studied, allowing for easier comparison with other research \cite{kn:Haider2020} \cite{kn:Garg2022} \cite{kn:Zhai2016}.

The primary focus of our study was the clustering of data based on two key variables: Dimensions and Severity. We assessed the quality of the resulting clusters using the Silhouette Score proposed by Rousseeuw (\cite{kn:Rousseeuw1987}). To assess the stability and robustness of our clustering results, we conducted a sensitivity analysis. Specifically, we varied parameters for all clustering approaches. For instance, we performed a sensitivity analysis on the K-means algorithm by varying the number of clusters (k) in a predefined range, i.e. from 2 to 15. This sensitivity analysis allowed us to evaluate how the clustering results vary with different parameter settings and identify an optimal number of clusters. To determine the membership of each observation in a specific dimension, we adopted two approaches. The first approach relied on identifying the most frequent dimension within the cluster to which the observation belongs. The second approach considered the share of the observations in the cluster relative to the total share of the same dimension cases within the cluster. These approaches allowed us to assign observations to the appropriate dimensions based on their clustering characteristics.

We calculate accuracy as the ratio of correctly assigned data points to the total number of data points in the dimension. We repeat this process for each cluster and dimension, and finally, we compute the total accuracy for each dimension across all clusters. This allows for easy comparison and analysis of accuracy across different clustering methods, numbers of clusters, and dimensions.

\subsection{Forecasting}
XGBoost (Extreme Gradient Boosting) is an enhanced version of the GBDT (Gradient Boosting Decision Tree) model. It aims to combine multiple decision trees with lower accuracy to create a model with higher accuracy \cite{kn:Chen2016}. The XGBoost algorithm incorporates the concept of gradient descent in each tree's generation. It iteratively moves towards the direction of minimizing the objective function based on the previously generated tree. By iterating through multiple decision trees, the algorithm continually reduces the loss error, resulting in the final prediction model. The split nodes of each decision tree follow the CART (Classification and Regression Tree) criteria, commonly utilizing the least square loss and logarithmic function. The algorithm's specific principles are as follows:

Let ${(x_i, y_i) | 1 \leq i \leq n}$ represent a given dataset with $n$ samples and $m$ features. Here, $x_i$ is the feature vector corresponding to sample $i$, and $y_i$ is the true value of the target category. XGBoost employs regression trees as the base classifier and predicts the output using the sum of $K$ CART trees. The input value to the function $f_k(x)$ is the error between the predicted value $\hat{y}_i^{(k-1)}$ and the true value, which reduces the loss function. The prediction model of XGBoost is defined by formula:

\begin{eqnarray}
\hat{y}_i = \sum_{k=1}^{K} f_k(x_i; \Theta),
\end{eqnarray}

\noindent where $\Theta$ represents the set of parameters.

XGBoost determines the complexity of each tree using the function $f$ defined in formula:

\begin{eqnarray}
f = \frac{1}{2} \sum_{j=1}^{T} \left(\frac{{\sum_{i \in I_j} w_i}}{{\sum_{i \in I_j} w_i + \lambda}} + \gamma\right) + \lambda T,
\end{eqnarray}

\noindent where $I_j$ represents the leaf nodes, $T$ is the number of leaf nodes, $w_i$ denotes the weight value of each leaf node, $\lambda$ is the regularization coefficient, and $\gamma$ represents the difficulty of node segmentation. The lower the value of $f$, the lower the complexity of the tree, resulting in better generalization.

The target loss function is defined by formula:

\begin{eqnarray}
L = \sum_{i=1}^{n} l(y_i, \hat{y}_i) + \sum_{k=1}^{K} \Omega(f_k),
\end{eqnarray}

\noindent where $l(y, \hat{y})$ represents the loss function between the true value $y$ and the predicted value $\hat{y}$. The training error is given by $\sum_{i=1}^{n} l(y_i, \hat{y}_i)$. XGBoost further approximates the loss function using the second-order Taylor expansion, resulting in the loss function shown in formula:

\begin{eqnarray}
L \approx \sum_{i=1}^{n} \left[g_i f(x_i) + \frac{1}{2} h_i f^2(x_i)\right] + \Omega(f),
\end{eqnarray}

\noindent where $g_i$ and $h_i$ are the first and second derivatives of the loss function, respectively.

The formula for $g_i$ and $h_i$ is given by:

\begin{eqnarray}
g_i = \frac{\partial l(y_i, \hat{y}_i)}{\partial \hat{y}_i}, \quad h_i = \frac{\partial^2 l(y_i, \hat{y}_i)}{\partial \hat{y}_i^2}
\end{eqnarray}

Let ${x_i \vert 1 \leq i \leq n, x_i \in \mathbb{R}^d}$ represent the sample set of leaf nodes with index $j$. The mapping of the sample space to the category space is denoted as $o: \{1, 2, 3, \ldots\} \rightarrow \{1, 2, 3, \ldots\}$. From formula (5) and the definition of $Z_j$, where $Z_j$ represents the modulus of the leaf node vector, we can transform formula (7) into the following form:

\begin{eqnarray}
L \approx \sum_{j=1}^{T} \left[\frac{G_j^2}{H_j + \lambda} + \gamma\right] + \lambda T + \Omega(f),
\end{eqnarray}

\noindent where $G_j$ and $H_j$ are defined as follows:

\begin{eqnarray}
G_j = \sum_{i \in I_j} g_i, \quad H_j = \sum_{i \in I_j} h_i
\end{eqnarray}

The optimal solution $Z_j^*$ is obtained using maximum likelihood estimation and can be determined as:

\begin{eqnarray}
Z_j^* = -\frac{G_j}{H_j + \lambda}
\end{eqnarray}

Therefore, the optimal solution of the objective function can be obtained as shown in formula:

\begin{eqnarray}
L^* = -\frac{1}{2} \sum_{j=1}^{T} \frac{G_j^2}{H_j + \lambda} - \gamma T
\end{eqnarray}

Logistic regression, an extensively utilized regression technique for forecasting the anticipated outcome of a binary dependent variable, was employed as a benchmark in this study. The logistic regression model,
as a component of the broader generalized linear models framework, is defined by a specific collection of predictor variables \cite{kn:McCullagh1989}. 

\subsection{Model evaluation}

The evaluation of the model was conducted using a confusion matrix based on the test data \cite{kn:Powers2011}. The confusion matrix provides valuable information such as precision, recall, F1 score, and accuracy.

Precision and recall are calculation metrics used to assess the effectiveness of information retrieval \cite{kn:Manning2009}. Precision measures the accuracy of the system in classifying text as either hoax or valid. The precision formula is defined as follows:
\begin{eqnarray}
Precision = TP / (TP + FP),
\end{eqnarray}
\noindent where TP is the number of True Positives and FP is the number of False Positives.

Recall, also known as sensitivity or true positive rate, measures the system's ability to identify all relevant instances. Recall is calculated using the formula:
\begin{eqnarray}
Recall = TP / (TP + FN), 
\end{eqnarray}
\noindent where FN is the number of False Negatives and TP is the number of True Positives.

F1-score is a metric that combines precision and recall into a single value, providing a balanced measure of performance. It is calculated using the following formula:
\begin{eqnarray}
F1-score = 2 * ((Precision * Recall) / (Precision + Recall))
\end{eqnarray}

Accuracy is used to evaluate the overall correctness of the system's classification. The accuracy formula is as follows:
\begin{eqnarray}
Accuracy = (TP + TN) / (TP + FP + TN + FN) * 100
\end{eqnarray}
\noindent where TP is the number of True Positives, FP is the number of False Positives, TN is the number of True Negatives, and FN is the number of False Negatives.

\section{Empirical results}

\subsection{Clustering analyses}

The purpose of our clustering analysis was to determine whether certain validation dimensions exhibit distinct characteristics that enable their identification through clustering methods. Specifically, we aimed to assess whether observations from specific dimensions could be captured by clusters that contain a high proportion of these dimensions, indicating their approximation. Conversely, we investigated whether certain dimensions were not easily distinguishable and thus could result in clusters containing observations from multiple dimensions.

We conducted the clustering analysis on findings, considering varying numbers of allowed clusters ranging from 2 to 15. The minimum number of clusters was set to 2, while the maximum was set to 15 to avoid inconclusive results that may arise from an excessive number of clusters.

As anticipated, the predictability of finding validation dimensions increased with the number of clusters utilized, allowing for the identification and separation of more distinct characteristics (Table 1). For 15 or fewer clusters, the overall prediction rate reached approximately 60\%, with certain dimensions being more readily identifiable than others. The best result was achieved by Birch algorithm.

\begin{table}
\begin{center}
{\caption{Comparison of accuracy for various clustering algorithms. Note: AC means Agglomerative Clustering, SC - Spectral Clustering}\label{table1}}
\begin{tabular}{lccccc}
\hline
\rule{0pt}{12pt}
No. of Clusters & AC & Birch & KMeans & Mini-Batch & SC \\
\hline
2 & 36.1\% & 35.6\% & 35.6\% & 27.5\% & 27.5\% \\
3 & 44.0\% & 45.2\% & 46.7\% & 41.9\% & 28.5\% \\
4 & 44.0\% & 46.9\% & 45.1\% & 37.6\% & 34.4\% \\
5 & 46.0\% & 48.6\% & 47.9\% & 40.5\% & 34.4\% \\
6 & 46.0\% & 48.6\% & 47.0\% & 47.2\% & 31.1\% \\
7 & 46.0\% & 51.3\% & 46.6\% & 45.4\% & 37.7\% \\
8 & 49.0\% & 52.4\% & 53.1\% & 45.8\% & 39.9\% \\
9 & 50.7\% & 52.4\% & 51.3\% & 47.2\% & 39.4\% \\
10 & 50.7\% & 52.8\% & 51.1\% & 50.7\% & 40.0\% \\
11 & 53.6\% & 53.6\% & 49.8\% & 52.1\% & 41.9\% \\
12 & 53.6\% & 56.2\% & 54.0\% & 53.6\% & 40.3\% \\
13 & 54.2\% & 56.8\% & 57.1\% & 53.6\% & 44.9\% \\
14 & 54.6\% & 56.8\% & 57.5\% & 52.4\% & 42.3\% \\
15 & 54.9\% & 58.1\% & 56.9\% & 57.5\% & 46.1\% \\
\hline
\end{tabular}
\end{center}
\end{table}

When employing the approach that assigns the predicted dimension based on the most frequent dimension within the cluster to which an observation belongs, the dimensions of impact assessment, margin of conservatism, model design, and model input achieved predictability rates above 60\% (Table 2). However, for some dimensions like Model Environment, Model Implementation, and Model Use, the accuracy was always 0\%, indicating that this method is probably not appropriate for underrepresented dimensions.

\begin{table}
\begin{center}
{\caption{Predictability of validation dimension using the most frequent cases in clusters as predictors (in \%) Note: C means Number of clusters, Doc - Documentation, Imp - Impact Assessment, MoC - Margin of Conservatism, Des - Model Design, Env - Model Environment, Impl - Model Implementation, In - Model input, Out - Model output, Use - Model Use}\label{table2}}
\begin{tabular}{l|cccccccccccccc}
\hline
\rule{0pt}{12pt}
C & Doc & Imp & MoC & Des & Env & Impl & In & Out & Use \\
\hline
\\[-6pt]
2 & 0.0 & 0.0 & 76.8 & 0.0 & 0.0 & 0.0 & 100.0 & 0.0 & 0.0 \\
3 & 0.0 & 0.0 & 76.8 & 91.4 & 0.0 & 0.0 & 64.6 & 0.0 & 0.0\\
4 & 0.0 & 0.0 & 76.8 & 33.1 & 0.0 & 0.0 & 64.6 & 89.3 & 0.0\\
5 & 0.0 & 0.0 & 76.8 & 64.0 & 0.0 & 0.0 & 64.6 & 58.3 & 0.0\\
6 & 0.0 & 0.0 & 76.8 & 64.0 & 0.0 & 0.0 & 64.6 & 58.3 & 0.0\\
7 & 0.0 & 85.7 & 76.8 & 64.0 & 0.0 & 0.0 & 64.6 & 58.3 & 0.0\\
8 & 0.0 & 85.7 & 76.8 & 63.3 & 0.0 & 0.0 & 69.1 & 58.3 & 0.0\\
9 & 0.0 & 85.7 & 76.8 & 63.3 & 0.0 & 0.0 & 69.1 & 58.3 & 0.0\\
10 & 80.8 & 85.7 & 76.8 & 63.3 & 0.0 & 0.0 & 47.5 & 58.3 & 0.0\\
11 & 80.8 & 85.7 & 76.8 & 41.7 & 0.0 & 0.0 & 66.9 & 58.3 & 0.0\\
12 & 40.4 & 85.7 & 76.8 & 41.7 & 0.0 & 0.0 & 87.8 & 58.3 & 0.0\\
13 & 40.4 & 85.7 & 76.8 & 39.6 & 0.0 & 0.0 & 87.8 & 65.0 & 0.0\\
14 & 40.4 & 85.7 & 76.8 & 39.6 & 0.0 & 0.0 & 87.8 & 65.0 & 0.0\\
15 & 40.4 & 85.7 & 76.8 & 66.9 & 0.0 & 0.0 & 87.8 & 36.9 & 0.0\\
\hline
\end{tabular}
\end{center}
\end{table}

Interestingly, the alternative prediction method, which randomly selects validation dimensions based on their share within the cluster of a specific observation, yielded less precise results (Table 3). This method accurately predicted only two dimensions, Impact Assessment, and Margin of Conservatism, with rates exceeding 60\% for 15 clusters.  The accuracy was also close to 60\% for Model Input. In this case accuracy for Model Environment, Model Implementation, and Model Use was also small (around 14-20\%) but was significantly higher than zero.

\begin{table}
\centering
\caption{Predictability of validation dimension using shares of findings from a specific dimension in clusters as predictors (in \%). Note: C means Number of clusters, Doc - Documentation, Imp - Impact Assessment, MoC - Margin of Conservatism, Des - Model Design, Env - Model Environment, Impl - Model Implementation, In - Model input, Out - Model output, Use - Model Use}
\label{table3}
\begin{tabular}{l|cccccccccccccc}
\hline
\rule{0pt}{12pt}
C & Doc & Imp & MoC & Des & Env & Impl & In & Out & Use \\
\hline
2 & 8.4 & 3.5 & 73.3 & 23.1 & 7.2 & 4.8 & 30.1 & 16.8 & 3.0\\
3 & 20.5 & 5.1 & 73.5 & 31.4 & 8.1 & 9.7 & 39.5 & 26.4 & 3.4\\
4 & 20.5 & 7.7 & 73.6 & 31.5 & 9.4 & 9.7 & 42.4 & 33.9 & 5.2\\
5 & 20.5 & 12.1 & 74.4 & 31.5 & 9.9 & 9.7 & 42.4 & 38.2 & 5.2\\
6 & 23.7 & 12.1 & 74.4 & 31.7 & 10.0 & 13.9 & 49.2 & 38.2 & 6.3\\
7 & 23.8 & 85.9 & 74.7 & 33.3 & 10.4 & 13.9 & 49.3 & 39.4 & 6.3\\
8 & 23.8 & 85.9 & 74.7 & 33.7 & 10.9 & 13.9 & 50.9 & 39.4 & 6.5\\
9 & 23.8 & 85.9 & 74.7 & 35.5 & 10.9 & 13.9 & 51.1 & 44.7 & 6.5\\
10 & 30.1 & 86.0 & 74.7 & 35.6 & 13.6 & 14.0 & 51.3 & 44.7 & 7.7\\
11 & 30.1 & 86.0 & 74.7 & 40.3 & 14.6 & 14.0 & 52.1 & 44.8 & 8.2\\
12 & 30.9 & 86.0 & 74.8 & 40.3 & 14.6 & 20.7 & 56.7 & 44.8 & 14.8\\
13 & 30.9 & 86.0 & 74.8 & 41.0 & 14.7 & 20.7 & 57.1 & 45.1 & 14.8\\
14 & 31.0 & 86.0 & 74.8 & 41.0 & 14.7 & 20.7 & 57.2 & 45.1 & 14.8\\
15 & 31.0 & 86.1 & 74.8 & 42.4 & 14.7 & 20.7 & 57.3 & 49.6 & 14.8\\
\hline
\end{tabular}
\end{table}

Generally, both prediction methods succeeded in forecasting Impact Assessment, and Margin of Conservatism but failed to accurately predict the dimensions of model environment, model implementation, and model use in the findings.

\subsection{Predictive power of individual tokens}

In our initial analysis, we assessed the predictive power of individual tokens (i.e., words) present in the validation findings. To this end, logistic regression and XGBoost models were employed, with each token serving as a potential predictor for a validation dimension. Notably, the XGBoost algorithm demonstrated superior performance in terms of precision. Consequently, we present the top 10 valuable predictors for each validation dimension derived from this model in Table 4 \footnote {Note: MoC denotes margin of conservatism – a special multiplier increasing the calculated PD, LGD, and EAD due to uncertainty present in the model; MD denotes a model development team, ISD is the implementation specification document describing technical specification of the investigated model. In the model implementation, one country name is mentioned, where some loan portfolios of the bank are present.}. Certain words, such as "document," "documentation," and "description" for the model documentation dimension, exhibited self-explanatory associations.

\begin{table}
\begin{center}
{\caption{Top 10 words predicting a validation dimension.}\label{table4}}
\begin{tabular}{lccccccc}
\hline
\rule{0pt}{12pt}
\quad Model Dimension & & Words \\
\hline
\rule{0pt}{12pt}
\\[-6pt]
\quad Model documentation& & model, data, document, code, \\
\quad & & description, default, level, \\
\quad & & order, documentation, update \\
\hline
\rule{0pt}{12pt}
\quad Model input& & model, use, customer, data, \\
\quad & & default, development, \\
\quad & & portfolio, MoC, perform, LGD\\
\hline
\rule{0pt}{12pt}
\quad Model design& &  use, model, MD, MoC, \\
\quad & & default, data, downturn, \\
\quad & & LGD, process, risk\\
\hline
\rule{0pt}{12pt}
\quad Model output& & data, LGD, observe, \\
\quad & & threshold, rating, perform, \\
\quad & & model, grade, test, month\\
\hline
\rule{0pt}{12pt}
\quad Margin of Conservatism & & MoC, analysis, \\
\quad & & development, deficiency, apply, \\
\quad & & default, MD, model, driver, perform \\
\hline
\rule{0pt}{12pt}
\quad Impact assessment& & impact, calculate, \\
\quad & & model, analysis, different, \\
\quad & & development, description, definition, \\
\quad & & default, deficiency\\
\hline
\rule{0pt}{12pt}
\quad Model implementation& & model, ISD, perform, \\
\quad & & implementation, implement, \\
\quad & & [country name here], include,\\
\quad & & plan, scope, update\\
\hline
\rule{0pt}{12pt}
\quad Model environment& & portfolio, development, \\
\quad & & model, scope, \\
\quad & & representativeness, risk, analysis, \\
\quad & & include, within, credit\\
\hline
\rule{0pt}{12pt}
\quad Model use& & use, model, decision, new, \\
\quad & & process, data, customer, \\
\quad & & credit, need, update\\
\hline
\\[-6pt]
\end{tabular}
\end{center}
\end{table} 

Likewise, we identified "perform" for the model output dimension, "MoC" for the margin of conservatism, "impact" and "calculate" for the model impact dimension, "ISD" and "implementation" for the model implementation dimension, and "scope" and "representativeness" for the model environment dimension. Additionally, "use," "process," and "decision" were determined as significant predictors for the model use dimension. Interestingly, logistic regression yielded very similar results to XGBoost.

However, the discriminatory power of models based solely on individual tokens was found to be far from perfect. The validation dimensions exhibited satisfactory prediction within the training sample, the quality of predictions in the testing and validation samples was limited. All three statistics—precision, recall, and F1-score—indicated discriminatory power below 80\% for certain dimensions, occasionally even dropping below 50\%. These findings underscore that treating validation findings as isolated words without considering the complete contextual information hinders the attainment of perfect predictions for validation dimensions.

\subsection{Integrating textual context into predictive models}

In order to enhance our analysis, we incorporated numerical vectors representing phrase embeddings as additional predictive factors in our model for explaining finding dimensions. The XGBoost modeling algorithm was employed for this purpose, and the list of predictive factors was expanded to include phrase embeddings, title words, and finding words.

The dataset was divided into three subsets: a training set comprising 60\% of all observations, a validation sample consisting of 20\% randomly selected observations, and a testing sample containing the remaining 20\% of observations. The analysis commenced by training the model with initial hyperparameter values. We then optimized the predictions by adjusting the hyperparameter values and evaluating the predictive power in the validation sample. Finally, we assessed the model's quality by analyzing its predictive performance in the testing sample.

The final predictive power achieved a value of 1.000 in the training sample, and slightly lower values of 0.781 and 0.785 were observed in the validation and testing samples, respectively (Table 5, Table 6, and Table 7). This represents a significant improvement over the specification of the XGBoost model. The model demonstrated precision above 80\% in predicting the dimensions of model input, output, implementation, design, environment, and impact assessment. However, the dimensions of model documentation and use exhibited precision below 70\%, as indicated by the F1-score statistic.

\begin{table}
\begin{center}
{\caption{Classification Metrics for train sample}\label{table5}}
\begin{tabular}{lccccccc}
\hline
\rule{0pt}{12pt}
Dimension&Precision&Recall&F1-Score&Support\\
\hline
\\[-6pt]
\quad documentation&1&1&1&29\\
\quad Impact assessment&1&1&1&12\\
\quad margin of conservatism&1&1&1&46\\
\quad model design&1&1&1&84\\
\quad model environment&1&1&1&29\\
\quad model implementation&1&1&1&15\\
\quad model input&1&1&1&102\\
\quad model output&1&1&1&65\\
\quad model use&1&1&1&12\\
\quad accuracy&&&1&394\\
\quad macro avg&1&1&1&394\\
\quad weighted avg&1&1&1&394\\
\hline
\end{tabular}
\end{center}
\end{table}

\begin{table}
\begin{center}
{\caption{Classification Metrics for Validation sample}\label{table6}}
\begin{tabular}{lccccccc}
\hline
\rule{0pt}{12pt}
Dimension&Precision&Recall&F1-Score&Support\\
\hline
\\[-6pt]
\quad documentation&1&0.79&0.88&14\\
\quad Impact assessment&1&1&1&3\\
\quad margin of conservatism&0.93&0.93&0.93&14\\
\quad model design&0.71&0.81&0.76&31\\
\quad model environment&1&0.5&0.67&8\\
\quad model implementation&0.9&0.9&0.9&10\\
\quad model input&0.85&0.82&0.84&34\\
\quad model output&0.75&0.94&0.83&16\\
\quad model use&0.5&0.5&0.5&2\\
\quad accuracy&&&0.83&132\\
\quad macro avg&0.85&0.8&0.81&132\\
\quad weighted avg&0.84&0.83&0.82&132\\
\hline
\end{tabular}
\end{center}
\end{table}

\begin{table}
\begin{center}
{\caption{Classification Metrics for test sample}\label{table7}}
\begin{tabular}{lccccccc}
\hline
\rule{0pt}{12pt}
Dimension&Precision&Recall&F1-Score&Support\\
\hline
\\[-6pt]
\quad documentation & 0.8 & 0.44 & 0.57 & 9\\
\quad Impact assessment&1&0.67&0.8&6\\
\quad margin of conservatism&0.64&1&0.78&9\\
\quad model design&0.69&0.83&0.75&24\\
\quad model environment&1&0.88&0.93&8\\
\quad model implementation&1&0.5&0.67&4\\
\quad model input&0.89&0.89&0.89&45\\
\quad model output&0.74&0.77&0.76&22\\
\quad model use&1&0.5&0.67&4\\
\quad accuracy &&&0.8&131\\
\quad macro avg&0.86&0.72&0.76&131\\
\quad weighted avg&0.82&0.8&0.8&131\\
\hline
\end{tabular}
\end{center}
\end{table}

These less favorable results suggest that certain flaws in model documentation cannot be easily distinguished from other deficiencies, particularly when the respective modeling issues raised by the validation team are inadequately described in the documentation. Furthermore, the dimension of model use may be closely related to model design, such as when predictive variables of an IRB model cannot be easily employed in credit decision-making, and to model documentation, such as when potential applications of the IRB model are not thoroughly described.

To interpret the contribution of each explanatory variable to the final prediction, we utilized SHAP (SHapley Additive exPlanations) values. The contribution of an explanatory variable is quantified as the average difference between the predicted value of the target variable in the presence of the particular explanatory variable and the predicted value without it. Results presented in Figure 2 confirm that phrase embeddings play a central role in assigning findings to specific dimensions, surpassing the significance of individual words.

\begin{figure}
\centerline{\includegraphics[height=3.5in]{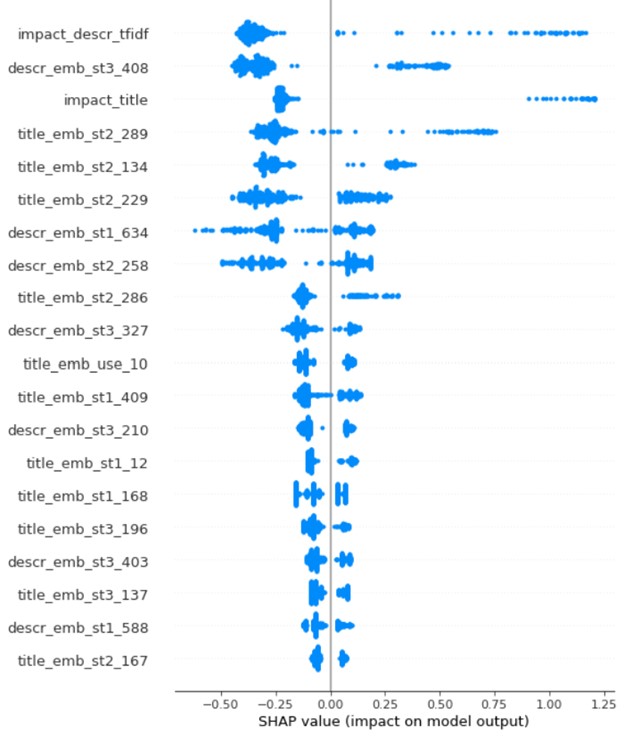}}
\caption{Contributions of individual words and word embeddings on model predictions in the model identifying finding dimensions} \label{procstructfig}
\end{figure}

\section{Conclusion}
In conclusion, our empirical analysis aimed to explore the predictability of validation dimensions in the context of clustering analyses and machine learning techniques. We utilized logistic regression and XGBoosting models to assess the predictive power of individual tokens extracted from validation findings. While these models showed some success in predicting validation dimensions in the training sample, their performance was limited in the testing and validation samples, indicating that individual words alone do not capture the full context of findings for accurate predictions.

To overcome this limitation, we incorporated clustering analysis into our study, aiming to identify distinct clusters that approximate specific validation dimensions. The predictability of finding validation dimensions increased as the number of clusters used increased. Notably, the approach of assigning the predicted dimension based on the most frequent dimension in the cluster yielded more precise results compared to the alternative method of randomly selecting validation dimensions based on their share in the cluster.

To further enhance the predictive models, we integrated phrase embeddings representing the textual context of findings into the predictive factors. This addition significantly improved the overall predictive power, particularly in terms of precision. The model achieved a predictive power of 1.000 in the training sample and demonstrated a precision above 80\% for dimensions such as model input, output, implementation, design, environment, and impact assessment. However, the precision for model documentation and use dimensions remained below 70\%, suggesting challenges in distinguishing certain deficiencies and issues related to model documentation and use.

The analysis of SHAP values revealed that phrase embeddings played a crucial role in allocating findings to specific dimensions, emphasizing their significance in the prediction process.

Overall, our findings highlight the importance of considering the contextual information encoded in phrase embeddings for accurate predictions of validation dimensions. This approach provides valuable insights for improving the understanding and assessment of validation findings in various domains. Future research could explore additional methods to further enhance the precision of predictions and expand the application of these techniques in practical settings.

\bibliography{ecai}
\end{document}